%% file: main.tex
\documentclass{ecai} 

\usepackage{latexsym}
\usepackage{amssymb}
\usepackage{amsmath}
\usepackage{amsthm}
\usepackage{booktabs}
\usepackage{enumitem}
\usepackage{graphicx}
\usepackage{color}
\usepackage{multirow}
\usepackage{float}

\newcommand{\BibTeX}{B\kern-.05em{\sc i\kern-.025em b}\kern-.08em\TeX}

\begin{document}

%%%%%%%%%%%%%%%%%%%%%%%%%%%%%%%%%%%%%%%%%%%%%%%%%%%%%%%%%%%%%%%%%%%%%%%%

\begin{frontmatter}

\paperid{6616} 

%%% Use this command to specify the title of your paper.

\title{Probability Density from Latent Diffusion Models for Out-of-Distribution Detection}

%%% Use this combinations of commands to specify all authors of your 
%%% paper. Use \fnms{} and \snm{} to indicate everyone's first names 
%%% and surname. This will help the publisher with indexing the 
%%% proceedings. Please use a reasonable approximation in case your 
%%% name does not neatly split into "first names" and "surname".
%%% Specifying your ORCID digital identifier is optional. 
%%% Use the \thanks{} command to indicate one or more corresponding 
%%% authors and their email address(es). If so desired, you can specify
%%% author contributions using the \footnote{} command.

\author[A]{\fnms{Joonas}~\snm{Järve}\thanks{Corresponding Author. Email: joonas.jarve@ut.ee}}
\author[A]{\fnms{Karl Kaspar}~\snm{Haavel}}
\author[A]{\fnms{Meelis}~\snm{Kull}} 

\address[A]{Institute of Computer Science, University of Tartu, Estonia}

%%% Use this environment to include an abstract of your paper.

\begin{abstract}
Despite rapid advances in AI, safety remains the main bottleneck to deploying machine‑learning systems. A critical safety component is out‑of‑distribution (OOD) detection: given an input, decide whether it comes from the same distribution as the training data. In generative models, the most natural OOD score is the data likelihood. Actually, under the assumption of uniformly distributed OOD data, the likelihood is even the optimal OOD detector, as we show in this work.
However, earlier work reported that likelihood often fails in practice, raising doubts about its usefulness. 
We explore if, in practice, the representation space also suffers from the inability to learn good density estimation for OOD detection, or if it is merely a problem of the pixel-space typically used in generative models. To test this, we trained a Variational Diffusion Model (VDM) not on images, but on the representation space of a pre-trained ResNet-18 to assess the performance of our likelihood-based-detector in comparison to the state-of-the-art methods from OpenOOD suite.
\end{abstract}

\end{frontmatter}

%%%%%%%%%%%%%%%%%%%%%%%%%%%%%%%%%%%%%%%%%%%%%%%%%%%%%%%%%%%%%%%%%%%%%%%%

\section{Introduction}
\label{sec:intro}

Out-of-distribution detection is essential for the reliable and safe operation of AI systems that use machine learning.
For safety-critical applications, machine learning models must be able to assess whether input data is novel or different from training data. However, both generative deep neural networks and discriminative classifiers can fail in distinguishing between in-distribution (InD) data and OOD \cite{nalisnickDeepGenerativeModels2019,choi_waic_2018,nguyen2015deep}. Furthermore, \citet{fangOutDistributionDetectionLearnable} proved an impossibility theorem, demonstrating that universal OOD-detection is not perfectly learnable from in-distribution only. 

Existing OOD detection methods can be divided into three main categories \cite{yang_generalized_2021}: classification-based, density-based, and distance-based methods. Classification-based approaches rely on confidence scores produced by classification models, treating high-confidence inputs as InD and low-confidence inputs as OOD. Density-based methods use probabilistic models to capture the distribution of the data and use likelihoods as scores, which are most commonly estimated in pixel space. However, earlier work has shown that likelihood often fails in OOD detection by assigning higher likelihoods to OOD than InD inputs. This behaviour occurs because density-based methods often prioritise low-level statistical features, and in high dimensions, the highest likelihood regions may not contain InD data ~\cite{nalisnickDeepGenerativeModels2019, choi_waic_2018}. Distance-based methods measure the distance between an input and the closest point in the training data in feature space to distinguish between InD and OOD samples. 
Many of the ``non-density'' methods can be thought of as proxy density estimators, for example, Mahalanobis OOD detector \cite{Lee2018}, Energy-based OOD detector \cite{liu2021energybasedoutofdistributiondetection} and KNN based detection \cite{sun2022outofdistributiondetectiondeepnearest} because they assess how isolated a point is in feature space -- larger distances imply lower density, mimicking the behaviour of density estimators.
Since many methods effectively approximate density estimation, it stresses the necessity to shed more light into the methods that directly model data density and density-based OOD detection.

\begin{figure}[ht]
\centering
\includegraphics[width=0.9\linewidth]{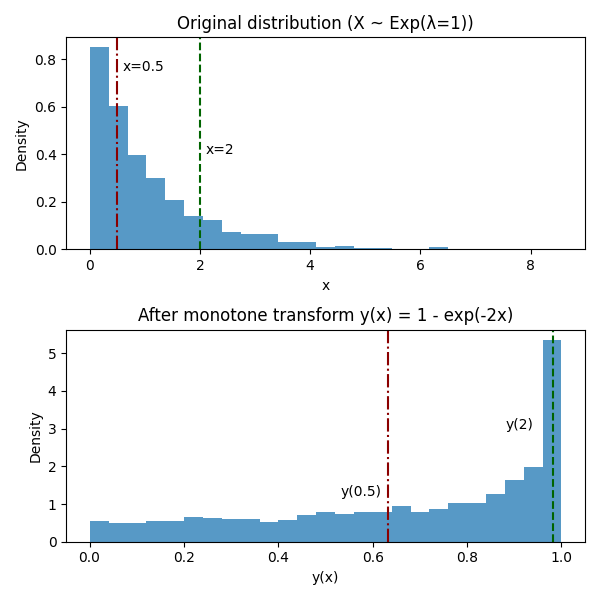}
\caption{Illustration of monotonic transformation on the samples that alters the density such that the point $x = 0.5$ with higher density ends up with lower density after transformation than point $x=2$ with initially lower density.  }
\label{fig:exmp1}
\end{figure}

Density estimation can be carried out in a variety of representational spaces — whether in the original input domain, in a learned latent space, or even after applying simple monotone transformations — and each choice of space can fundamentally alter the shape of the underlying probability distribution. Even a single probability distribution may exhibit very different high- and low-density regions depending on how it is reparametrized, i.e., depending on the metric or coordinates in which it is evaluated. In Figure \ref{fig:exmp1}, we illustrate that a point which originally lies in the most concentrated region can, after a basic monotone transformation, be mapped into what appears to be a low-density area; conversely, points from sparse regions may end up in regions of high estimated density. Thus, density-based OOD detection is inherently space-dependent. This makes density-based OOD detection non-obvious and raises the question of whether densities can consistently give good OOD detection performance and in what conditions they are optimal. In Section \ref{sec:motivation}, we show that, under the assumption of a uniform OOD distribution, an OOD detector based solely on in-distribution data density is ROC/AUC--optimal, justifying the use of densities for OOD detection.\\

It has been shown that generative models trained in pixel space \cite{nalisnickDeepGenerativeModels2019,kirichenkoWhyNormalizingFlows2020,nalisnick_detecting_2019} cannot provide reliable likelihoods to detect OOD. We hypothesize that this failure is specific to the pixel space, and density estimation in the latent space works better. Hereafter, by the term ``latent space representations'' or ``latent representations'', we mean the same as hidden representations, penultimate layer activations, feature embeddings and use them interchangeably throughout the paper. To test the hypothesis, we train a Variational Diffusion Model (VDM) \cite{kingma_variational_2023} to estimate the density in the representation space as they claim state-of-the-art density estimation. Using a strong estimator avoids confounding results due to suboptimal density-estimator.
While VDMs yield only a lower bound on log-likelihood, exact log-likelihoods can be obtained via their score using probability-flow ODEs \cite{songScoreBasedGenerativeModeling2021}. Using the latent representations of an encoder to estimate density makes it into sort-of-a Latent Diffusion Model (LDM) \cite{rombachHighResolutionImageSynthesis2022}, but without a decoder. We provide a detailed comparison of a VDM-based OOD detector on the OpenOOD benchmark \cite{zhangOpenOODV15Enhanced2023}, comparing it to state-of-the-art (SOTA) OOD detection methods. However, our goal is not to claim superior performance, but to show that density-based methods in representation space can be competitive and merit further study.
\newline

Our main contributions are as follows:
\begin{enumerate}
    \item Under the assumption of a uniform OOD distribution, an OOD detector that relies exclusively on in-distribution density achieves the optimal AUC.
    \item We demonstrate that exact log-likelihoods from a well-trained latent VDM match SOTA OOD detection performance—implying the latent space to be fit for OOD detection based on the densities.
    \item In addition to the likelihood-based, we also propose a diffusion-loss-based OOD detection method that proves more robust and stable in benchmarks.
    \item We provide a thorough evaluation of our VDM-based OOD detectors on the OpenOOD benchmark \cite{zhangOpenOODV15Enhanced2023}, benchmarking them against leading methods.
\end{enumerate}
%%%%%%%%%%%%%%%%%%%%%%%%%%%%%%%%%%%%%%%%%%%%%%%%%%%%%%%%%%%%%%%%%%%%%%%%

%-------------------------------------------------------------------------
\section{Related work}\label{sec:related_lit}

Denoising diffusion probabilistic models (DDPMs) were introduced by \citet{ho_denoising_2020}  for high-quality image reconstruction. DDPMs define a fixed forward noising schedule and train a network to reverse it by denoising, thereby sampling from the learned distribution. DDPMs were expanded with variational diffusion models (VDMs), simplifying the variational lower bound (VLB), optimizing the noise schedule, and introducing a continuous-time diffusion process. Furthermore, DDPMs have been extended to accommodate higher resolutions \cite{rombachHighResolutionImageSynthesis2022}. 

Multiple methods have been proposed to use diffusion models for OOD detection. \citet{graham_denoising_2022} used DDPMs for OOD detection by measuring reconstruction errors across various noise levels, hypothesising that InD images exhibit consistent similarity, while OOD samples show significant divergence. OOD score was based on a combination of MSE and LPIPS, to measure the distance between the original and reconstructed image.

DDPMs are mostly used for reconstruction-based OOD detection, but they can be used to calculate likelihood as done by \citet{goodier_likelihood-based_2023}. They proposed an OOD score called complexity corrected likelihood ratio (CCLR). CCLR takes into account that generative models trained on InD data can assign higher likelihood to a subset of OOD data \cite{nalisnick_detecting_2019,choi_waic_2018}. This is due to the fact that likelihoods produced by generative models are skewed by low-level features that generalise across datasets, overshadowing the high-level, unique semantics crucial for OOD detection. To achieve robust likelihood-based OOD detection, it is necessary to correct for this low-level image complexity \cite{goodier_likelihood-based_2023}. \citet{hengOutDistributionDetectionSingle2024} introduce DiffPath, showing that simple geometric cues—the rate‑of‑change and curvature of the probability‑flow ODE trajectory—obtained from a pixel‑space unconditional model already separate InD from OOD samples without retraining. Most recently, \citet{zhou_nodi_2024} exploit the noise vectors predicted by a diffusion model—and even an analytic closed‑form surrogate—to build encoder‑agnostic OOD scores that remain robust across CNN and ViT backbones. 
 
However, most diffusion-model-based methods add noise to the inputs and measure OOD scores in the pixel space, rather than applying noise directly within the representation space. Our method departs from these pixel‑space formulations by learning a variational diffusion process directly in an encoder’s latent manifold.

Density estimation in the feature space has been proposed in several nondiffusion-model-related OOD detection works already. However, many of them \cite{liuSimplePrincipledUncertainty2020, mukhotiDeepDeterministicUncertainty2023,amersfoortUncertaintyEstimationUsing2020,  caoDeepHybridModels2022} have been using a \emph{Spectral Normalization} (SN) \cite{miyatoSpectralNormalizationGenerative2018} in the encoder. Using SN can confound attribution in the sense that it can achieve better OOD detection not due to better density estimation but better representation \cite{liuSimplePrincipledUncertainty2020}. Others \cite{cook2024featuredensityestimationoutofdistribution,erdil2021taskagnosticoutofdistributiondetectionusing,sun2022outofdistributiondetectiondeepnearest} lack a comparison to the state-of-the-art methods given in the OpenOOD benchmark \cite{zhangOpenOODV15Enhanced2023}.

We address these gaps and learn a density estimator VDM in the non-spectral-normalized representation space. We will also compare the performance of likelihood-based scores to the noise-vector-based scores in the OOD detection task.

\section{Motivation}\label{sec:motivation}

Recent advances in the use of likelihood-based OOD detection and density‐estimation techniques naturally raise a foundational question of whether, even in principle, density estimates can reliably distinguish between InD and OOD samples. 

A key challenge in this domain is the intrinsic uncertainty regarding where, in the feature space, OOD examples will manifest. If one has prior knowledge of plausible OOD regions, then strategies such as outlier exposure \cite{Hendrycks2018} can be employed to substantially enhance detection performance. However, in the absence of such foreknowledge, a possible position is to assume a uniform distribution over the OOD domain. We investigate this uniformity assumption, examining its theoretical implications and describing how—and to what extent—it affects the ability of density estimates to flag OOD observations. Building on the work of \citet{bishop1993novelty}, if the novel class is uniform on the region of input space (and zero elsewhere), the Bayes decision reduces to thresholding the InD density. We sharpen and extend this: under the same assumption, \emph{thresholding the InD density is also ROC/AUC--optimal}, because it induces the same ranking as the likelihood ratio.
\newline

Let's have instance space $\mathcal{X}\subset\mathbb{R}^d$ and label space $\mathcal{Y} = \{1,\dots,K\}$, where $K\in \mathbb{N}$. Now, let's denote the InD distribution as $P_I(X_I,Y_I)\sim \mathcal{X}\times\mathcal{Y}$, where random vectors $X_I\in \mathcal{X}$ and $Y_I \in \mathcal{Y}$. Similarly, let's denote the OOD distribution $P_O(X_O)$ with random vector $X_O\in \mathcal{X}$ but $P_O\ne P_I$. The OOD samples may come from unknown classes $\mathcal{Y'}$, with $\mathcal{Y'}\cap\mathcal{Y}=\emptyset$.

Let the random label $Y\in\{I,O\}$ indicate whether a sample $X\in\mathbb R^d$ is InD or OOD.
Let's assume the joint mixture
\[
P(X,Y)
=(1-\alpha)\,P_I(X)\,\mathbf 1_{\{Y=I\}}
\;+\;
\alpha\,P_O(X)\,\mathbf 1_{\{Y=O\}},
\]
where $0<\alpha<1.$ Hence the class--conditional densities are
\[
P(X\mid Y=I)=P_I(X),\qquad 
P(X\mid Y=O)=P_O(X),
\]
with priors $P(Y=I)=1-\alpha$ and $P(Y=O)=\alpha$.

Bayes' rule gives the posterior
\begin{align}
\pi_O(X):=P(Y=O\mid X)=&\frac{P(X\mid Y=O)\cdot P(Y=O)}
{P(X)}= \nonumber \\ 
=&\frac{\alpha\,P_O(X)}
{\alpha\,P_O(X)+(1-\alpha)\,P_I(X)}.
\label{eq:posterior}
\end{align}

Let's recall that Neyman--Pearson lemma \cite{1933RSPTA.231..289N} states that the likelihood--ratio statistic
\begin{equation}
r(X)=\frac{P_O(X)}{P_I(X)}    
\label{eg:rx}
\end{equation}
maximises the true--positive rate at every false--positive rate. 
Since AUC depends only on the score ordering, any strictly increasing transform of $r(X)$—including $\pi_O(X)$ from \eqref{eq:posterior}—is ROC/AUC-optimal. Indeed, \eqref{eq:posterior} rewrites as
\[
\pi_O(X)=\frac{1}
{1+\dfrac{1-\alpha}{\alpha}\,\dfrac{1}{r(X)}},
\]
which is strictly increasing in $r(X)$.

A natural way for an OOD detector to declare a sample OOD would be whenever $P_I(X)$ is ``sufficiently small''.  
Therefore, let the score be a strictly decreasing function of the InD density:
\[
s_I(X)=f\bigl(P_I(X)\bigr).
\]
For $s_I$ to be the ROC/AUC--optimal, it must be a monotone (here decreasing) function of the likelihood ratio $r(X)$; that is,
\begin{equation}
r(X)=g\bigl(P_I(X)\bigr)
\qquad \forall X,
\label{eq:monotone_requirement}
\end{equation}
for some \emph{scalar} function $g$.

Rewriting \eqref{eg:rx} as
\begin{equation}
P_O(X)=P_I(X)\,g\bigl(P_I(X)\bigr),
\label{eq:Po_uniform}
\end{equation}
the right--hand side depends only on the value $s=P_I(X)$, not on $X$ itself.  
Consequently, every level set of $P_I$ (all $X$ such that $P_I(X)=s$) must have the same $P_O$ value.  
One such density satisfying this for all level sets is a constant: $P_O(X)=c, \, \forall X$ and $c>0$
i.e.\ $P_O$ is uniform on the support where $P_I>0$. Then, $r(x)\propto 1/P_I(X)$, hence any decreasing function $f$ of $P_I$ induces the same ranking as $r$ and is ROC/AUC--optimal.
\newline

We demonstrated that there exist representation spaces where density-based OOD detection achieves optimal AUC, justifying the use of densities in principle. This supports the idea that density estimation is not fundamentally flawed in OOD detection, and can, in fact, theoretically yield optimal OOD detection performance when using densities in such a space where OOD data are uniform. Besides, we hypothesize that it's always possible to transform the space such that the uniformity assumption holds, but this will be left as future work.

A notable method that seemingly leverages this result is \citet{hendrycks_deep_2018}. They introduce a methodology of incorporating an auxiliary dataset of known outliers to training by encouraging them to be detected approximately as uniform distribution.

%%%%%%%%%%%%%%%%%%%%%%%%%%%%%%%%%%%%%%%%%%%%%%%%%%%%%%%%%%%%%%%%%%%%%%%%
\section{Methodology}

In this section, we will go over the key ingredients of our methodology -- from hidden representations to exact likelihoods.

Let's denote the training dataset by $\mathcal{D} = \{(x_i, y_i)\}_{i=1}^n$ drawn i.i.d from  $P_I(X_I,Y_I)$.
A deterministic encoder $\phi : \mathbb{R}^{d}\to\mathbb{R}^{m}$ maps inputs to latents \( \mathcal{Z}=\{\phi(x): (x,y)\in \mathcal{D}\}\) that are used to train the diffusion model in our work. 
\newline

In the following paragraphs, we will go over the pre-existing theory of VDMs~\cite{kingma_variational_2023} and \emph{Probability-flows} to explain how our density estimates are formulated. Since VDMs just provide a lower bound for the log-likelihood, then using its score in a probability-flow ODE, we can arrive at the exact log-likelihoods. As in the previous literature, we have not explicitly seen how to adapt VDM into a probability-flow ODE, we spend a little more space on explaining it.

\subsection{Variational Diffusion Models}
Following \citet{kingma_variational_2023}, we are using continuous--time VDMs, where both gradual forward diffusion and denoising process timesteps $t\sim \mathcal{U}(0,1)$ are sampled from a uniform distribution. The forward process defines noisy versions of the input $z\in\mathcal{Z}$, which are called latent variables $z_t$. Time $z_0$ means noise-free input, and $z_1$ means the most noisy input.  
The noising process is given as follows:
$$q(z_t|z) \sim \mathcal{N}( \alpha_tz,\sigma_t^2I), \quad z_t = \alpha_tz + \sigma_t\epsilon_t,\quad \epsilon_t \sim\mathcal{N}(0,I). $$
Since we are using a variance-preserving process, then $\alpha_t^2 = 1 - \sigma_t^2$ and the noise schedule is parametrized $\sigma_t^2 = \operatorname{sigmoid}(\gamma_{\mathbf\eta}(t))$, where $\gamma_{\eta}(t)$ is learnable monotonic neural network, parametrised by $\eta$ as proposed by \cite{kingma_variational_2023}.  

The denoising process is given by $$q(z_s| z_t,z) \sim \mathcal{N}(\mu_Q(z_t,z;s,t), \sigma_Q^2(s,t)I),$$ where $\mu_Q(z_t,z;s,t)$ and $\sigma_Q^2(s,t)$ are parameterized by the noise predicting network, and $s$ and $t$ are two consecutive timesteps, $t>s$.

We remind that VDMs are trained by optimising the \emph{variational lower bound} (VLB):
$$ VLB(z) = \mathbb{E}_{q(z_0|z)}\big[ \log p(z|z_0) \big] - D_{KL}(q(z_1|z)||p(z_1)) - \mathcal{L}_\infty(z) $$
which provides us a lower bound on likelihood as well.
$$ \log{p(z)} \geq VLB(z).$$
We omit the first term of VLB in our training because by construction $z_0\approx z$ as explained in \cite{ribeiroDemystifyingVariationalDiffusion2024}; this is a common simplification in VDM training. The diffusion loss $\mathcal{L}_{\infty}$ is a MSE between predicted noise and actual noise:
\begin{equation}\label{eq:diff_loss}
    \mathcal{L}_{\infty}(z) = \frac 12 \mathbb{E}_{\epsilon\sim\mathcal{N}(0,I),t\sim U(0,1)}\bigg[|| \epsilon -\hat \epsilon_{\theta}(z_t,t)||_2^2\bigg].
\end{equation}

\subsection{Probability--Flow ODE}
Following ~\cite{songScoreBasedGenerativeModeling2021}, we can think of VDM as modeling a denoising process from the prior distribution $p_T(z)$ to data distribution $p_0(z)$ over time $t$ as Stochastic Differential Equation (SDE):
\[
\mathrm dz
  = \mathbf f( z,t)\,\mathrm dt
  + g(t)\,\mathrm d\mathbf w,
\]
where \(\mathbf f( z,t)\) is the drift, \(g(t)\ge 0\) the scalar
diffusion coefficient, and \(\mathbf w\) a standard Wiener process.

The SDE can be expressed as a reverse-time SDE as follows:
$$dz = [\mathbf{f}(z,t) - g(t)^2\nabla\log p_t(z)]dt + g(t)d\mathbf{\hat w}$$

Any diffusion model that predicts either the noise or the score can be interpreted as the gradient of the log-density \(\nabla_{z}\log p_t(z)\) of a reverse--time SDE~\cite{songSlicedScoreMatching2019,songScoreBasedGenerativeModeling2021}.
Replacing the stochastic term with its mean yields the
\emph{probability--flow ODE}
\[
\frac{\mathrm dz}{\mathrm dt}
  = \mathbf f( z,t)
    -\frac12\,g(t)^{2}s_\theta( z_t,t)
\]
whose solution shares the same set of marginals \(\{p_t\}_{t\in[0,T]}\)
as the SDE.

As mentioned before, VDMs do not prescribe a
hand‑crafted diffusion coefficient \(g(t)\) like DDPMs; instead, they learn a monotone network
\(\gamma_{\eta}\colon[0,T]\!\to\!\mathbb R\) and set the noise
schedule $\sigma_t^{2} = \operatorname{sigmoid}(\gamma_{\eta}(t))$. Therefore, $g(t) = \sigma_t\sqrt{\gamma'_{\eta}(t)}$, which we have not seen formulated explicitly before. It is a key difference of using a VDM in probability-flow ODE, instead of a DDPM. Since our VDM is a noise predictor, then following \cite{ribeiroDemystifyingVariationalDiffusion2024}, we can express $s_\theta( z_t,t)=-\hat\varepsilon_{\theta}( z_t,t)/\sigma_t$. Substituting \(g(t)\) and $s_\theta(z_t,t)$ into the generic probability--flow ODE specializes it to VDM:
\begin{align*}
\frac{\mathrm d z}{\mathrm dt}
  &= -\frac12\,g(t)^2
    \bigl[ z + s_\theta( z_t,t)\bigr] =\\
    &= -\frac12\,\sigma_t\sqrt{\gamma'_{\eta}(t)}
    \bigl[z - \frac{\hat\varepsilon_{\theta}(z_t,t)}{\sigma_t}\bigr] = \mathbf{\hat f}_\theta(z_t,t).
\end{align*}

At \(t=T\), the flow terminates in an isotropic Gaussian
\(p_T(z)=\mathcal N\!\bigl(\mathbf 0,\sigma_T^{2}\mathbf I\bigr)\),
and the exact log‑likelihood is obtained from
\begin{equation}\label{eq:exact-likelihood}
\log p_0( z(0))
  = \log p_T(z(T))
    +\int_{0}^{T}
        \nabla\,
        \mathbf{\hat f}_\theta(z(t),t)
      \,\mathrm dt,    
\end{equation}
where importantly $z(t)\neq z_t$, since probability flow arrives at a deterministic solution, unlike the diffusion process. Equation \ref{eq:exact-likelihood} is computed in practice by using Skilling-Hutchinson~\cite{Skilling1989,hutch} trace estimator and in our case RK4 ODE Solver~\cite{RK4} similarly to \citet{songScoreBasedGenerativeModeling2021}. 

\subsection{Our Method}

Based on the previous description, we define three scores to detect OOD. 
\newline

\textbf{Exact Likelihood (EL)} method is the exact likelihood $\log p_0( z(0))$ from the Equation \ref{eq:exact-likelihood}. Note that, despite the name, it is exact up to desirable accuracy \cite{songScoreBasedGenerativeModeling2021}. This is the core method, since it will test directly how well densities from the latent VDM perform for the OOD detection. In the experiments, we denote this method as VDM $\log p_0(z)$.
\newline

\textbf{Prior Likelihood (PL)} method is to use the likelihood of the prior term $\log p_T( z(T))$ of the Equation \ref{eq:exact-likelihood}. If the score provided by the VDM is accurate, the flow should map an in‑distribution sample to a point \( z(T)\) that lies inside the typical set of the Gaussian prior \(p_{T}\). Presumably, for an OOD input, the estimated score is unreliable and the sample point would end up in a less dense area of the Gaussian. We denote this method as VDM $\log p_T(z)$.
\newline

\textbf{Top-$K$ Diffusion Loss (TKDL)} is a method leveraging the diffusion loss, given in Equation \ref{eq:diff_loss}, with fixed-time loss at $t=1$:
\begin{equation}\label{eq:diff_loss_topk}
\mathcal{L}_{diff} = \mathbb{E}_{\epsilon\sim\mathcal{N}(0,I)}\bigg[|| \epsilon -\hat \epsilon_{\theta}(z_1,1)||_2^2\bigg].    
\end{equation}
Since our VDM is trained with classifier-free guidance \cite{ho2022classifierfreediffusionguidance}, then we can condition the loss on the class. This method creates a vector of diffusion losses $\mathbf{\mathcal{L}_{diff}^K}$, where $K$ is the number of classes under consideration. The classes are the top-K predictions per sample from the base encoder $\phi$. Then, \emph{softmax} function is applied on the negated vector of $\mathbf{\mathcal{L}_{diff}^K}$ and final score is given as follows:
$$\mathrm{TKDL} = \max{\big(\mathrm{softmax}(\mathbf{-\mathcal{L}_{diff}^K})\big).}$$

In this method, we choose $t=1$, i.e., the noisiest timestep, since we saw from the likelihood-based methods EL and PL, that the differentiation is more robust close to the prior distribution. Additionally, we use the learning dynamics of the diffusion model as a heuristic for choosing $t=1$, as learning is typically faster at higher noise levels, therefore providing better separability between OOD and InD (diffusion loss depicted in various noise levels is given in the Appendix \ref{appendix:noise}).

Method TKDL is somewhat similar to NODI \cite{zhou_nodi_2024} as they exploited the analytical form of the predicted noise vectors for detection. We diverge from their work since we choose one specific noise level, sample it repeatedly, and use class-conditioning alongside. The same reasoning also holds with the similar work of \citet{graham_denoising_2022}, where they take advantage of a reconstruction error concatenated from the range of noise levels since reconstruction error is simplified into a noise error in VDMs, and have a direct relationship as shown in \cite{ribeiroDemystifyingVariationalDiffusion2024}. 

%%%%%%%%%%%%%%%%%%%%%%%%%%%%%%%%%%%%%%%%%%%%%%%%%%%%%%%%%

\begin{figure}[ht]
\centering
\includegraphics[width=0.99\columnwidth]{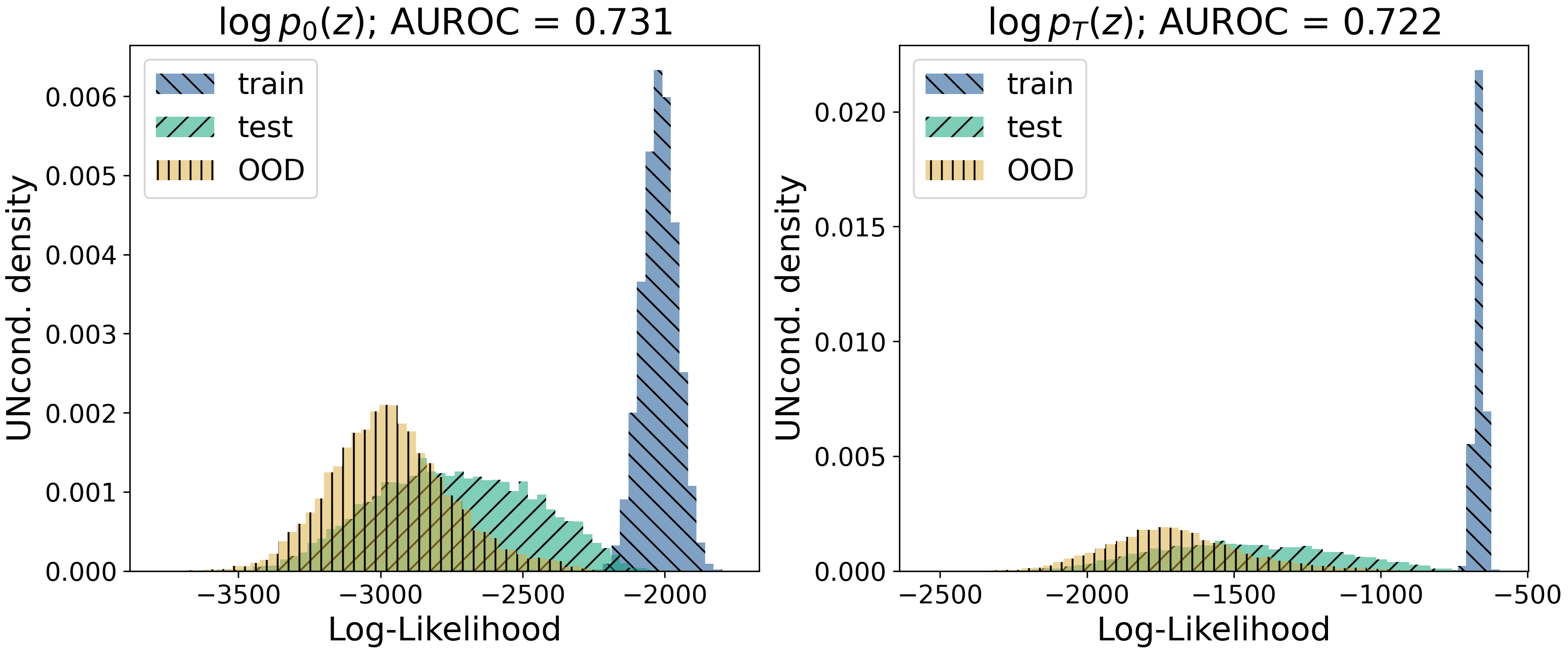}
\caption{Histogram of log‐likelihoods from a VDM trained on CIFAR-100 (epoch 1999): in‐distribution train (blue), test (green), and CIFAR-10 as OOD (orange). Left: $\log p_0(z)$ (AUROC : 0.731); right: $\log p_T(z)$ (AUROC : 0.722).}
\label{fig:dens_cif100}
\end{figure}

\begin{figure}[ht]
\centering
\includegraphics[width=0.99\columnwidth]{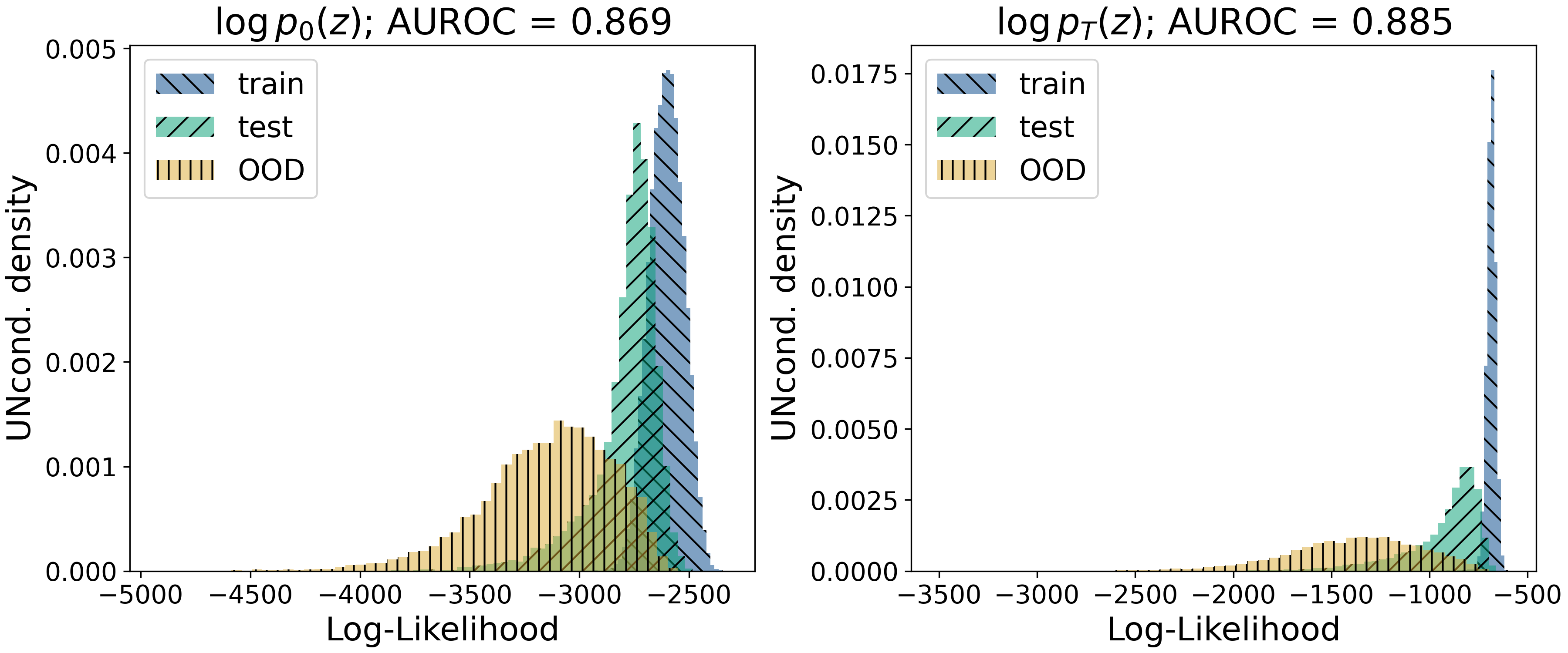}
\caption{Histogram of log‐likelihoods from a VDM trained on CIFAR-10 (epoch 1999): in‐distribution train (blue), test (green), and CIFAR-100 as OOD (orange). Left: $\log p_0(z)$ (AUROC : 0.869); right: $\log p_T(z)$ (AUROC : 0.885).}
\label{fig:dense_cif_10}
\end{figure}

\section{Experiments}

\subsection{Datasets and Models}

To get the most comprehensive understanding of the performance of our OOD detection methods, we used \emph{OpenOOD} \cite{zhangOpenOODV15Enhanced2023} benchmarks. All the encoders, datasets, data splits and model checkpoints were downloaded from their Github repository\footnote{https://github.com/Jingkang50/OpenOOD} to preserve the comparability with previous methods.

As an encoder, we used their pre-trained \emph{ResNet-18} \cite{resnet2015} models to get the hidden representations for our VDM. Prior to the training of the VDM, we first normalized the data to have zero mean and unit variance, ensuring the variance-preserving property. Additionally, we appended \emph{Fourier features} to the input, as suggested by the authors of VDM \cite{kingma_variational_2023}, to learn the fine details of the representations. We used the same two channels as the authors of ~\cite{kingma_variational_2023}, namely, $\sin(2^n\pi z)$ and $\cos(2^n\pi z)$, where $n=7$. The model training was classifier-free guided to be able to see the effect of class-conditioning in the detection methods. Since representations from the penultimate layer of an ResNet-18 are flat, we used a fully connected neural encoder-decoder network with residual connections and \emph{Layer Norm} \cite{ba2016layernormalization} as the backbone of the VDM. More details about the architecture of the VDM are given in Appendix \ref{appendix:vdm}. For training the VDM, we used a linear noise schedule as suggested in \cite{kingma_variational_2023} and all the models were trained for 2000 epochs. The TKDL method used top-5 classes from the encoder ResNet-18 and all the loss calculations were averaged over 20 times. Density estimates PL and EL used RK4 ODE solver from torchdiffeq \cite{torchdiffeq} library, integrating over 50 steps.
\newline

\textbf{Datasets.} To comply with OpenOOD benchmark datasets, we trained our VDM on CIFAR-10, CIFAR-100 \cite{cifar100} and ImageNet-200 \cite{5206848}. As OOD datasets, we used the OpenOOD defined hierarchy compiled of natural textural images of \emph{Texture} \cite{cimpoi2013describingtextureswild}, large-scale scene image dataset \emph{Places365} \cite{zhou2017places}, handwritten digit image dataset \emph{MNIST} \cite{mnist}, real-world house number digit dataset \emph{SVHN} \cite{svhn}, \emph{Tiny-ImageNet} \cite{tin}, large-scale species image dataset \emph{iNaturalist} \cite{inaturalist}, ImageNet OOD evaluation benchmark datasets \emph{NINCO} \cite{ninco} and \emph{OpenImage-O} \cite{wang2022vimoutofdistributionvirtuallogitmatching} and benchmark for semantic novelty detection \emph{SSB-Hard} \cite{ssbhard}. Some of those OOD datasets are considered Near-OOD and some Far-OOD \cite{detecting_ahmed,zhangOpenOODV15Enhanced2023}, depending on how different they are from InD dataset.
\newline

\textbf{Evaluation.} To evaluate the OOD detection, we use two of the most commonly used metrics. AUROC -- area under the receiver operating characteristic curve which plots the true positive rate against the false positive rate for each threshold level. The higher the AUROC, the better. The second metric we use for evaluation is FPR@95, which corresponds to a false positive rate when the true positive rate is 95\%. The smaller the FPR@95, the better.
\newline

\textbf{Baselines.} We compare our methods with 6 other methods that shared the same encoder and \emph{Cross Entropy} training method, did not use outlier data, and relied merely on the representations of the pre-trained ResNet-18, like us. We chose them based on their performance in OpenOOD benchmark rankings. Since there is no single best method across all three CIFAR-10, CIFAR-100, and ImageNet-200 benchmarks, we chose the \emph{best-performing} models in either Near-OOD or Far-OOD according to AUROC, to see how far our methods are from the SOTA. These 6 best-performing methods that we compare against are \emph{TempScale} \cite{tempscale}, \emph{RMDS} \cite{ren2021simplefixmahalanobisdistance}, \emph{MLS} \cite{hendrycks_scaling_2019}, \emph{VIM} \cite{wang2022vimoutofdistributionvirtuallogitmatching}, \emph{KNN} \cite{sun2022outofdistributiondetectiondeepnearest}, \emph{ASH} \cite{ash_ood}. 
\newline

\textbf{Computational Efficiency.} VDMs were trained for 2000 epochs on a single NVIDIA Tesla V100 (16GB VRAM), taking $\sim$200 minutes for CIFAR-10/100 and $\sim$1130 minutes for ImageNet-200. In contrast, simpler density-estimation baselines (KDE, Gaussian, GMM) covered in ablation, Section \ref{ablation}, fit in under a minute on an Intel i5-1135G7 CPU, except GMM, which takes a minute for CIFAR and $\sim$7.5 minutes for ImageNet-200.

\begin{table*}[ht]
    \centering
    \caption{AUROC ($\uparrow$) and FPR@95 ($\downarrow$) for each method across CIFAR-10, CIFAR-100 and ImageNet-200 under near- and far-OOD. \textbf{Bold} indicates the best result and \underline{underline} the second best per column.}
    \label{tab:performance_metrics}
    \resizebox{2.05\columnwidth}{!}{%
    \begin{tabular}{lccccccccccccc}
        \toprule
        & \multicolumn{6}{c}{Near-OOD} & \multicolumn{6}{c}{Far-OOD} \\
        \cmidrule(lr){2-7}\cmidrule(lr){8-13}
        & \multicolumn{2}{c}{CIFAR-10} & \multicolumn{2}{c}{CIFAR-100} & \multicolumn{2}{c}{ImageNet-200} 
        & \multicolumn{2}{c}{CIFAR-10} & \multicolumn{2}{c}{CIFAR-100} & \multicolumn{2}{c}{ImageNet-200} \\
        \cmidrule(lr){2-3}\cmidrule(lr){4-5}\cmidrule(lr){6-7}
        \cmidrule(lr){8-9}\cmidrule(lr){10-11}\cmidrule(lr){12-13}
        Method 
        & AUROC$\uparrow$ & FPR@95$\downarrow$ 
        & AUROC$\uparrow$ & FPR@95$\downarrow$ 
        & AUROC$\uparrow$ & FPR@95$\downarrow$ 
        & AUROC$\uparrow$ & FPR@95$\downarrow$ 
        & AUROC$\uparrow$ & FPR@95$\downarrow$ 
        & AUROC$\uparrow$ & FPR@95$\downarrow$ \\
        \midrule
        TempScale 
            &  88.09 & 50.96 & \underline{80.90} & \textbf{54.49} & \textbf{83.69} & \underline{54.82} 
            &  90.97 & 33.48 & 78.74 & 57.94 & 90.82 & 34.00 \\
        RMDS 
            & \underline{89.80} & \underline{38.89} & 80.15 & \underline{55.46} & 82.57 & \textbf{54.02}
            &  92.20 & 25.35 & \textbf{82.92} & \underline{52.81} & 88.06 & 32.45 \\
        MLS  
            &  87.52 & 61.32 & \textbf{81.05} & 55.47 & \underline{82.90} & 59.76
            &  91.10 & 41.68 & 79.67 & 56.73 & 91.11 & 34.03 \\
        VIM  
            &  88.68 & 44.84 & 74.98 & 62.63 & 78.68 & 59.19
            &  \textbf{93.48} & \underline{25.05} & 81.70 & \textbf{50.74} & 91.26 & \textbf{27.20} \\
        KNN  
            & \textbf{90.64} & \textbf{34.01} & 80.18 & 61.22 & 81.57 & 60.18
            &  \underline{92.96} & \textbf{24.27} & \underline{82.40} & 53.65 & \underline{93.16} & \underline{27.27} \\
        ASH  
            &  75.27 & 86.78 & 78.20 & 65.71 & 82.38 & 64.89
            &  78.49 & 79.03 & 80.58 & 59.20 & \textbf{93.90} & 27.29 \\
        VDM $\log p_0(z)$ 
            & 87.29 & 58.41 & 76.23 & 62.48 & 52.24& 93.11
            & 91.74 & 34.04 & 79.38& 53.79 & 63.83& 87.40 \\
        VDM $\log p_T(z)$ 
            &  89.36 & 40.19 & 75.87 & 64.46  & 59.09&90.71
            &  92.54& 26.83 & 78.05 & 56.60 &71.05& 82.69 \\
        TKDL 
            &  87.67 & 50.89 & 79.54 &55.75 &80.93 & 56.58
            & 90.39 & 35.99 & 76.76 &60.14 & 85.72& 41.53\\
        \bottomrule
    \end{tabular}}
\end{table*}
\subsection{Results}
In all of the following tables, \textbf{bold} refers to the best and \underline{underline} to the second best result in the benchmark.
\newline

The comparison of all the methods and datasets aggregated by the difficulty level, i.e., either Near-OOD or Far-OOD, is given in Table \ref{tab:performance_metrics}. We can see that across datasets and types of OOD, our only method that performs comparably to SOTAs is TKDL, still never outperforming them. Our density-based methods are matching performance in the CIFAR-10 dataset, both in Near- and Far-OOD cases. On the other hand, they are clearly underperforming in the ImageNet-200 benchmark. Since PL and EL are still performing relatively well on CIFAR-100, this gives reason to presume that the same VDM-based density estimator used for each training dataset might need more capacity to capture the 200-class dataset of ImageNet-200. 
\newline

From Table \ref{tab:combined_results_dif10}, we can see how our methods compare to baselines in the context of CIFAR-10 in AUROC and FPR@95 metrics. In addition, to show average performance among the methods, we added the \emph{average rank} (Avg Rk) metric as well. The average rank is the mean of both metrics' ordinal positions across all evaluated datasets, giving a single number that reflects their overall relative performance. We see from Table \ref{tab:combined_results_dif10} that our prior likelihood-based method is the second best in both the MNIST and Textures datasets regarding both metrics. Based on the average rank, the prior likelihood-based method is in the top 4 among all the methods. Similar tables for ImageNet-200 and CIFAR-100 are given in the Appendix \ref{appendix:experiments}. 
\newline

\begin{table*}[ht]
  \centering
  \caption{InD: CIFAR-10; AUROC($\uparrow$) and FPR@95\,(\%)($\downarrow$) scores and average rank for our methods and baselines. Best in each column in \textbf{bold}, second best \underline{underlined}.}
  \label{tab:combined_results_dif10}
  \resizebox{2.05\columnwidth}{!}{%
    \begin{tabular}{l*{7}{c}*{7}{c}}
      \toprule
      & \multicolumn{7}{c}{AUROC$\uparrow$} & \multicolumn{7}{c}{FPR@95\,(\%)$\downarrow$} \\
      \cmidrule(lr){2-8} \cmidrule(lr){9-15}
      Method & CIFAR-100 & TIN & MNIST & SVHN & Textures & Places365 & Avg Rk
             & CIFAR-100 & TIN & MNIST & SVHN & Textures & Places365 & Avg Rk \\
      \midrule
      TempScale                   & 87.17 & 89.00 & 93.11 & 91.66 & 90.01 & 89.11 & 6.00
                                  & 55.81 & 46.11 & 23.53 & 26.97 & 38.16 & 45.27 & 5.67 \\
      RMDS                        & \underline{88.83} & \underline{90.76} & 93.22 & 91.84 & 92.23 & \underline{91.51} & 3.50
                                  & \underline{43.86} & \underline{33.91} & 21.49 & 23.46 & 25.25 & \underline{31.20} & 2.83 \\
      MLS                         & 86.31 & 88.72 & 94.15 & 91.69 & 89.41 & 89.14 & 6.00
                                  & 66.59 & 56.06 & 25.06 & 35.09 & 51.73 & 54.84 & 7.50 \\
      VIM                         & 87.75 & 89.62 & \textbf{94.76} & \textbf{94.50} & \textbf{95.15} & 89.49 & \underline{2.50}
                                  & 49.19 & 40.49 & \textbf{18.36} & \textbf{19.29} & \textbf{21.14} & 41.43 & \underline{2.67} \\
      KNN                         & \textbf{89.73} & \textbf{91.56} & \underline{94.26} & 92.67 & 93.16 & \textbf{91.77} & \textbf{2.17}
                                  & \textbf{37.64} & \textbf{30.37} & \underline{20.05} & 22.60 & 24.06 & \textbf{30.38} & \textbf{1.83} \\
      ASH                         & 74.11 & 76.44 & 83.16 & 73.46 & 77.45 & 79.89 & 9.00
                                  & 87.31 & 86.25 & 70.00 & 83.64 & 84.59 & 77.89 & 9.00 \\
VDM $\log p_0(z)$                 & 86.45 & 88.14 & 92.11 & 93.28 & 93.44 & 88.11 & 6.00
                                  & 63.11 & 53.72 & 27.71 & 26.03 & 28.98 & 53.45 & 6.50 \\
VDM $\log p_T(z)$                 & 88.54 & 90.18 & 91.88 & \underline{93.92} & \underline{94.58} & 89.78 & 3.50
                                  & 44.14 & 36.23 & 26.11 & \underline{21.31} & \underline{22.70} & 37.19 & 3.33 \\
TKDL                              & 86.80 & 88.53 & 93.44 & 89.98 & 88.67 & 89.48 & 6.33
                                  & 54.18 & 47.61 & 22.56 & 37.35 & 44.21 & 39.85 & 5.67\\
      \bottomrule
    \end{tabular}%
  }
\end{table*}

\begin{figure}[ht]
\centering
\includegraphics[width=0.99\columnwidth]{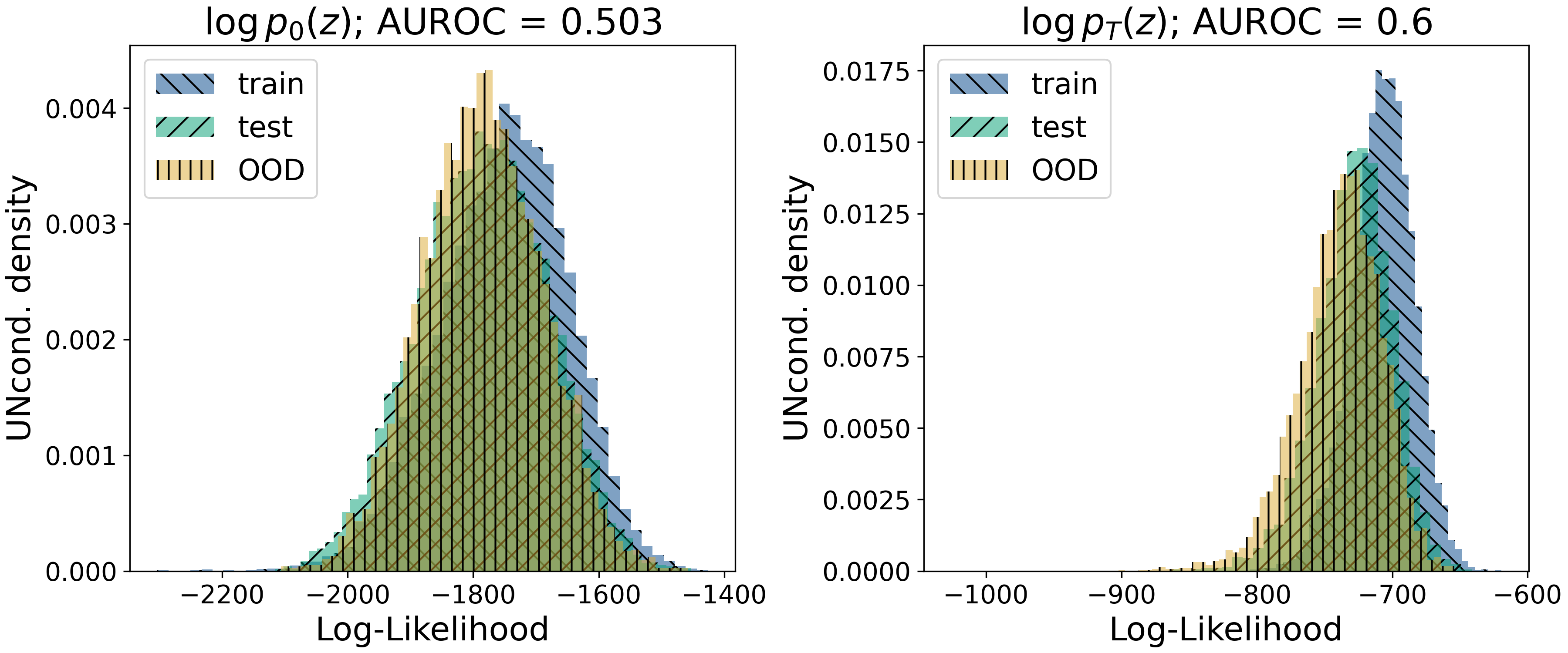}
\caption{Histogram of log‐likelihoods from a VDM trained on ImageNet-200 (epoch 1999): in‐distribution train (blue), test (green), and NINCO as OOD (orange). Left: $\log p_0(z)$ (AUROC : 0.503); right: $\log p_T(z)$ (AUROC : 0.6)}
\label{fig:dense_image}
\end{figure}

In Figures \ref{fig:dense_cif_10} and \ref{fig:dens_cif100}, we depict the traditional oppositions between CIFAR-10 and CIFAR-100. We see how the results of the tables are manifested in the histograms of log-likelihoods --- Figure \ref{fig:dense_cif_10} demonstrates a good separation between InD and OOD. In Figure \ref{fig:dens_cif100}, we see that the test data has diverged from the training distribution -- we can speculate this to be a result of an overfit. The latter combined with the ImageNet-200 \emph{versus} NINCO in Figure \ref{fig:dense_image}, we can see that densities tend to be more sensitive towards over- and underfit than the diffusion-loss based method TKDL.

\subsection{Conditional or Unconditional likelihoods?}
In Table \ref{tab:vdm_log_probs}, we can see the difference between class-conditional and unconditional density estimation across the benchmarks. In the case of CIFAR-10 and CIFAR-100, there are just minor differences between the conditional and unconditional likelihood estimation. In the case of ImageNet-200, the difference is more significant in both Near- and Far-OOD case. The latter phenomenon could be explained by the lack of capacity, possibly implying that longer training or a larger backbone would have been necessary for VDM, as it could not capture the detailed densities of all the 200 classes. Nevertheless, there does not seem to be any clear benefit in conditioning nor vice-versa in the context of likelihoods.

\begin{table}[ht]
    \centering
    \caption{AUROC$\uparrow$ of VDM log probabilities under different conditions across datasets.}
\resizebox{\columnwidth}{!}{%
    \begin{tabular}{lcccccc}
        \toprule
        & \multicolumn{2}{c}{CIFAR100} & \multicolumn{2}{c}{CIFAR10} & \multicolumn{2}{c}{ImageNet-200} \\
        \cmidrule(r){2-3} \cmidrule(r){4-5} \cmidrule(r){6-7}
        & Near-OOD & Far-OOD & Near-OOD & Far-OOD & Near-OOD & Far-OOD \\
        \midrule
        VDM $\log p_0(z)$ UNcond. & \underline{76.23} & \textbf{79.38} & 87.29 & 91.74 & 52.24 & 63.83 \\
        VDM $\log p_0(z)$ cond. & \textbf{76.25} & \underline{79.27} & 87.22 & 91.70 & 50.02 & 60.72 \\
        VDM $\log p_T(z)$ UNcond. & 75.87 & 78.05 & \textbf{89.36} & \textbf{92.54} & \textbf{59.09} & \textbf{71.05} \\
        VDM $\log p_T(z)$ cond. & 75.98 & 78.03 & \underline{89.31} & \underline{92.51} & \underline{56.18} & \underline{67.03} \\
        \bottomrule
    \end{tabular}}
    
    \label{tab:vdm_log_probs}
\end{table}

\subsection{Prior or Exact Likelihood?}
\begin{figure}[ht]
  \centering
  % first image
  \includegraphics[width=\linewidth
  ]{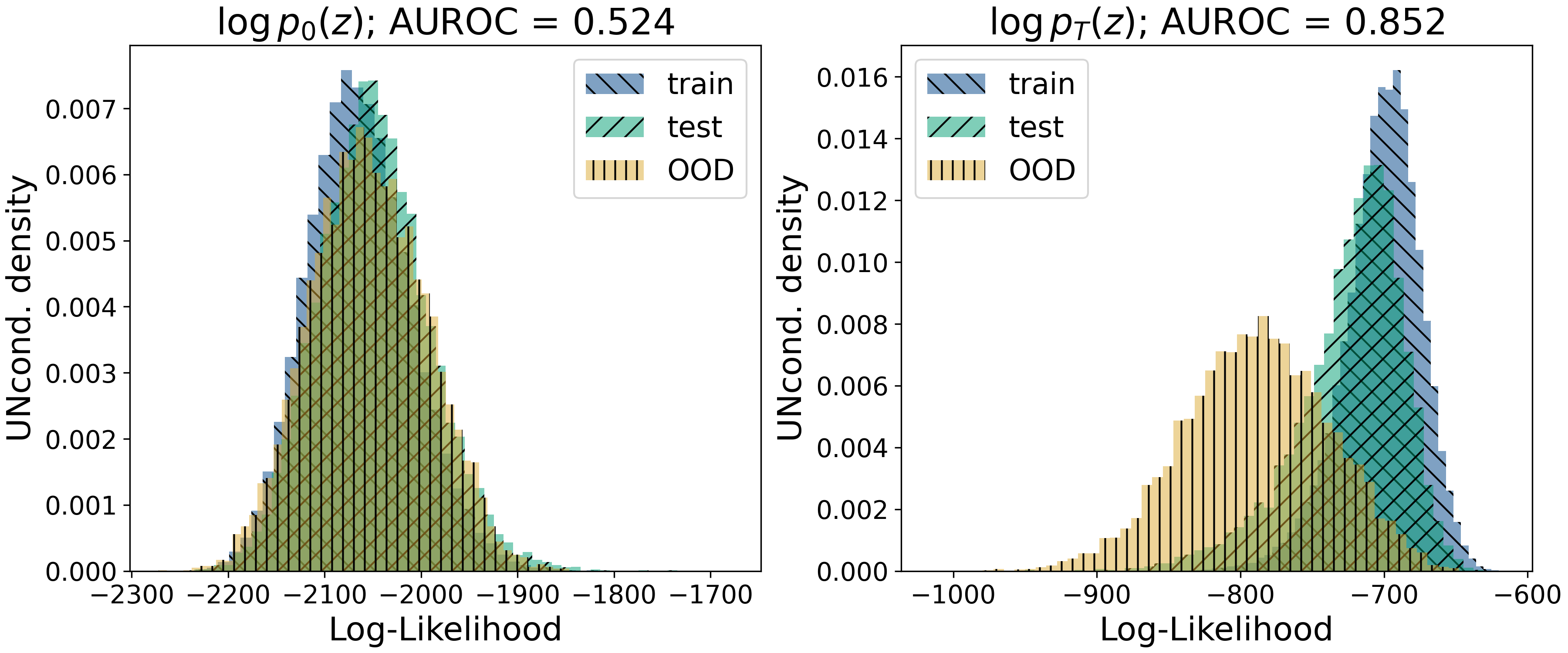}

  % small vertical gap (optional)
  \vspace{1em}

  % second image
  \includegraphics[
    width=\linewidth
  ]{LDM_ood_densities_cif10_cifar100_1999.png}

  \caption{Illustration of how \emph{test}, \emph{train}, and \emph{OOD} data densities look in different training phases. (top) On the left, there are depicted exact likelihoods and in the right, prior likelihoods after 59 epochs. (bottom)  On the left, there are depicted exact likelihoods and on the right, prior likelihoods after 1999 epochs.}
  \label{fig:epoch_diff}
\end{figure}

The question remains: which method should we prefer -- PL or EL, if it comes down to likelihoods? From Figure \ref{fig:epoch_diff}, we can see that training dynamics can have a say in it. In the top-left of Figure \ref{fig:epoch_diff}, at epoch 59, the exact likelihoods are inseparable, whereas in the top-right, at the same epoch, the OOD prior likelihoods already depart from test. In the bottom-left of the same figure, after 1999 epochs of training, separability has also become evident in the exact likelihood space. Hence, the use of prior likelihood can be justified and makes sense from the training dynamics as well, since the diffusion model tends to learn the noisier levels first.  

\subsection{Ablations with Simpler Density Estimators}\label{ablation}

We conducted additional ablations using simpler density-estimators: Kernel Density Estimation (KDE) \cite{kde_parzen,kde_rosenblatt}, Gaussian Mixture Model (GMM) and diagonal-covariance Gaussian, fitted on the same representations (details in Appendix \ref{appendix:baselines}). The AUROC scores, shown in Table \ref{tab:vdm_log_probsvsbaseline} (averaged over 3 runs as before), demonstrate that better density estimators provide better OOD detection performance. The only exception is the ImageNet-200 dataset, for which neither prior nor exact likelihood could outperform a simple Gaussian, possibly implying suboptimal diffusion model training or hyperparameters. However, all ablations perform worse than our TKDL.

\begin{table}[ht]
    \centering
    \caption{AUROC$\uparrow$ of VDM log probabilities compared to baseline density-estimators.}
\resizebox{\columnwidth}{!}{%
    \begin{tabular}{lcccccc}
        \toprule
        & \multicolumn{2}{c}{CIFAR100} & \multicolumn{2}{c}{CIFAR10} & \multicolumn{2}{c}{ImageNet-200} \\
        \cmidrule(r){2-3} \cmidrule(r){4-5} \cmidrule(r){6-7}
        & Near-OOD & Far-OOD & Near-OOD & Far-OOD & Near-OOD & Far-OOD \\
        \midrule
        VDM $\log p_0(z)$ UNcond. & \underline{76.23} & \textbf{79.38} & 87.29 & \underline{91.74} & 52.24 & 63.83 \\
        VDM $\log p_T(z)$ UNcond. & 75.87 & \underline{78.05} & \textbf{89.36} & \textbf{92.54} & 59.09 & 71.05 \\
        TKDL &  \textbf{79.54}  & 76.76 & \underline{87.67} &  90.39 & \textbf{80.93} & \textbf{85.72}  \\
        GMM & 63.58 & 75.08 & 62.90 & 64.84 & 64.85 & 65.77\\
        Gaussian & 70.48 & 67.21 & 69.24 & 75.4 & \underline{70.89} & \underline{77.19}\\
        KDE & 55.33 & 55.25 & 59.61 & 60.75 & 50.72 & 50.82\\
        \bottomrule
    \end{tabular}}
    
    \label{tab:vdm_log_probsvsbaseline}
\end{table}

%\paragraph{Limitations.}
%Our study relies on frozen ResNet-18 features and a single VDM backbone; likelihood quality depends on (i) representation quality and scaling, (ii) VDM capacity and training stability, and (iii) numerical tolerances in probability‑flow ODE integration and divergence estimation. Compute cost for EL/PL is higher than for simple geometric baselines.

\section{Conclusions and Future Work}

Generative models trained directly in pixel space are known to yield unreliable likelihoods for OOD detection \cite{nalisnickDeepGenerativeModels2019,kirichenkoWhyNormalizingFlows2020,nalisnick_detecting_2019}. We showed that density-based OOD detection is not intrinsically flawed, but its utility is space-dependent. Furthermore, we hypothesised that this limitation is specific to the input domain itself and that density estimation would be more effective in a representation space.  
To test this idea, we trained a Variational Diffusion Model (VDM) \cite{kingma_variational_2023} on hidden representations produced by a ResNet-18 encoder.
The combination of VDM with the probability‑flow ODE \cite{songScoreBasedGenerativeModeling2021} provides exact log‑likelihoods, not just a lower bound of log-likelihood, enabling a principled, density‑based OOD detector.
\newline

The comparison to the OpenOOD ResNet-18 benchmark showed that the VDM probability densities provide mostly comparable results to the leading OOD detection methods in the benchmark. Surprisingly, the Top-$K$ Diffusion Loss (TKDL) method provided even more robust and stable performance compared to the density-based approaches. Class-conditioning did not provide noticeable improvements to the likelihoods, but it made the TKDL method possible, enabling class-conditioned diffusion losses. We showed that by estimating density in the representation space, we can get OOD detection performance comparable to the state-of-the-art, implying the representation space to be fit for density-based OOD detection. 
\newline

As future work, different VDM backbones for better density estimates and stronger encoders for better representations could be tested. Similarly, evaluate whether spectral normalization improves representations; if so, extend it to generative models to stabilize their likelihood estimates.

%%%%%%%%%%%%%%%%%%%%%%%%%%%%%%%%%%%%%%%%%%%%%%%%%%%%%%%%%%%%%%%%%%%%%%%%

\begin{ack}
This work was supported by the Estonian Research Council grant PRG1604, by the Estonian Academy of Sciences research professorship in AI, and by the Estonian Centre of Excellence in Artificial Intelligence (EXAI), funded by the Estonian Ministry of Education and Research grant TK213.
\end{ack}

%%%%%%%%%%%%%%%%%%%%%%%%%%%%%%%%%%%%%%%%%%%%%%%%%%%%%%%%%%%%%%%%%%%%%%%%

%%% Use this command to include your bibliography file.

\bibliography{mybibfile}

%% use this to add an Appendix
\clearpage
\onecolumn
\appendix
\input{appendix.tex}

\end{document}

%% file: appendix.tex
\section{Training details}\label{appendix:vdm}
\subsection{VDM Architecture}
The backbone of the VDM is a fully connected encoder-decoder with residual connections. In the following, we give a list of specifications:
\begin{itemize}
    \item input dimensionality (hidden representation dim.): 512
    \item size of the time embedding: 128
    \item hidden dimensions of the encoder-decoder: {4096, 2048, 1024, 512, 256}
    \item class context vector size: 128
    \item activations used: GeLU \cite{hendrycks2023gaussianerrorlinearunits}
    \item normalization: Layer Normalization \cite{ba2016layernormalization}
\end{itemize}
The loss of the VDM was compromised by diffusion and latent loss, omitting reconstruction loss by construction of diffusion process. For training, we used batch size of 128, AdamW \cite{adamw} optimizer with weight decay of 0.01, learning rate of $0.0002$ with reducing learning rate on plateau\footnote{\url{https://docs.pytorch.org/docs/stable/generated/torch.optim.lr_scheduler.ReduceLROnPlateau.html}} (patience = 100, factor = 0.9).

\subsection{Density Baselines}\label{appendix:baselines}
For training the baseline density estimators, a hyperparameter search was carried out using $10000$ training samples and the rest of the training dataset as validation. In the case of Gaussian, we used an independent Gaussian, meaning a Gaussian with a diagonal covariance matrix. GMM and KDE were used from the \emph{scikit-learn} \cite{scikit-learn} library. For GMM, we tuned only the number of components (30, 20, and 50 components were used, respectively, for CIFAR-10, CIFAR-100, and ImageNet-200) and used the diagonal covariance matrix. For KDE, we tuned only the bandwidth size hyperparameter (0.1, 0.5, and 0.5 of bandwidth size were used, respectively, for CIFAR-10, CIFAR-100, and ImageNet-200) and the Gaussian kernel.

\section{Diffusion Loss over Different Noise Levels}\label{appendix:noise}

\begin{figure}[H]
\centering
\includegraphics[width=0.5\linewidth,trim=0cm 0cm 0cm 0cm,
    clip]{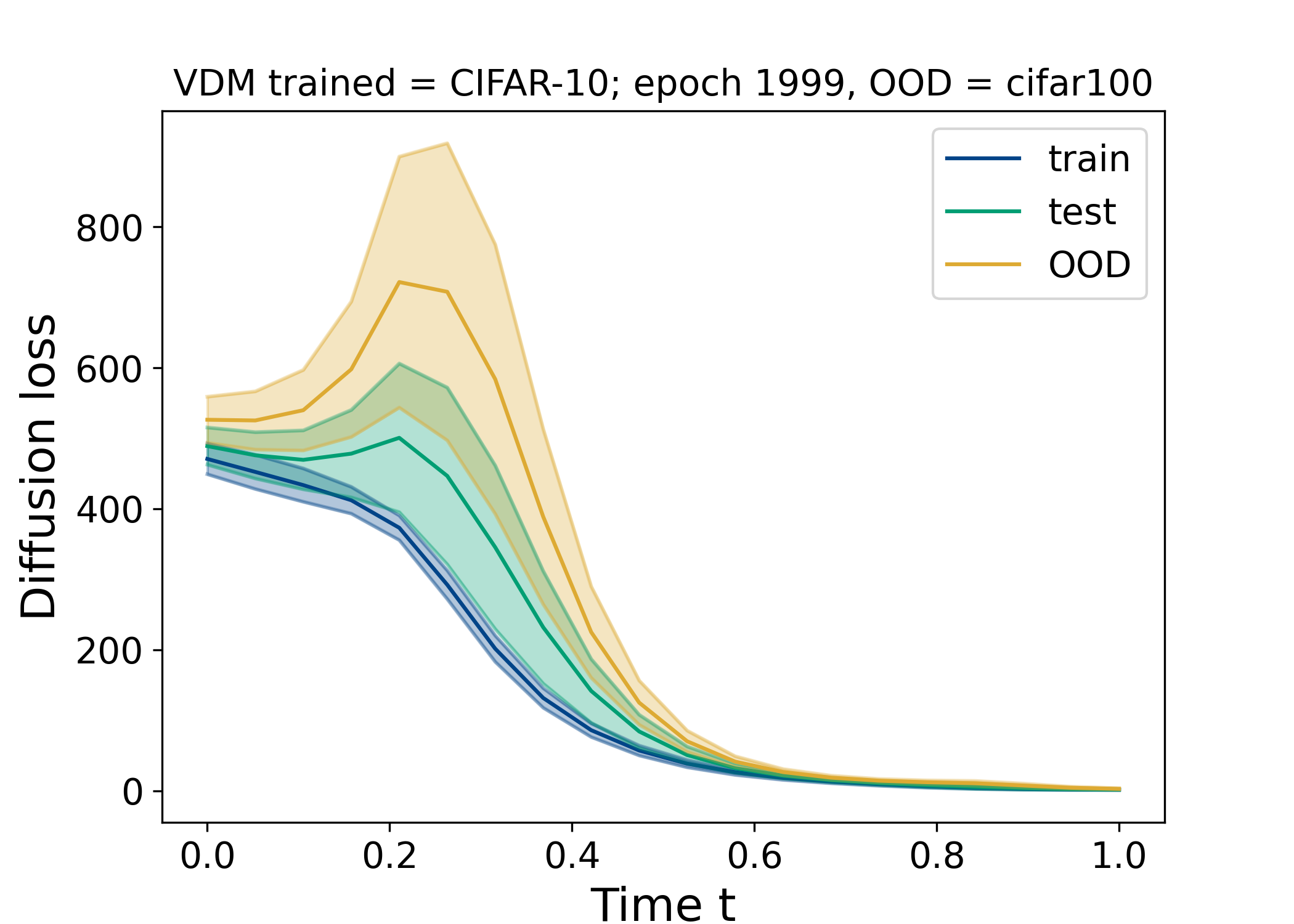}
\caption{Diffusion loss at different noise levels, given with variance.}
\label{fig:noise_schedule}
\end{figure}

\section{Additional Experiments}\label{appendix:experiments}
\subsection{CIFAR-10}

\begin{figure}[H]
\centering
\includegraphics[width=0.99\linewidth]{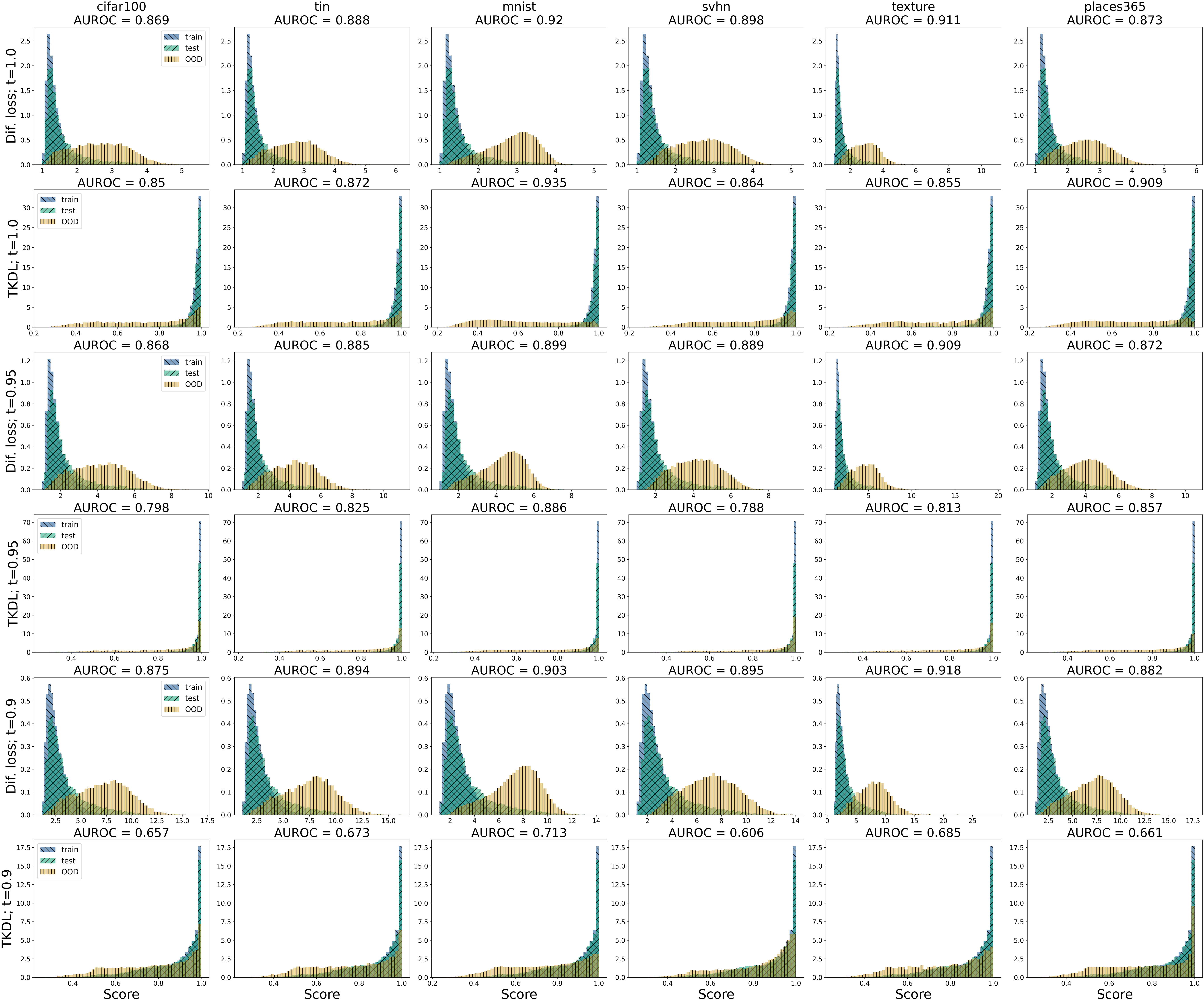}
    \caption{(InD: CIFAR-10) Comparison between using diffusion loss as an OOD score vs. our method TKDL with VDM trained for 2000 epochs. The resulting score histograms (InD train (blue), test (green), and OOD (orange)) and AUROC scores are given for each OOD dataset and for different diffusion timesteps (t=0.9, 0.95, 1.0) to demonstrate the relevance of the choice of timestep for both methods.}
\label{fig:basics_cif10_all}
\end{figure}

\begin{figure}[H]
\centering
\includegraphics[width=0.99\linewidth]{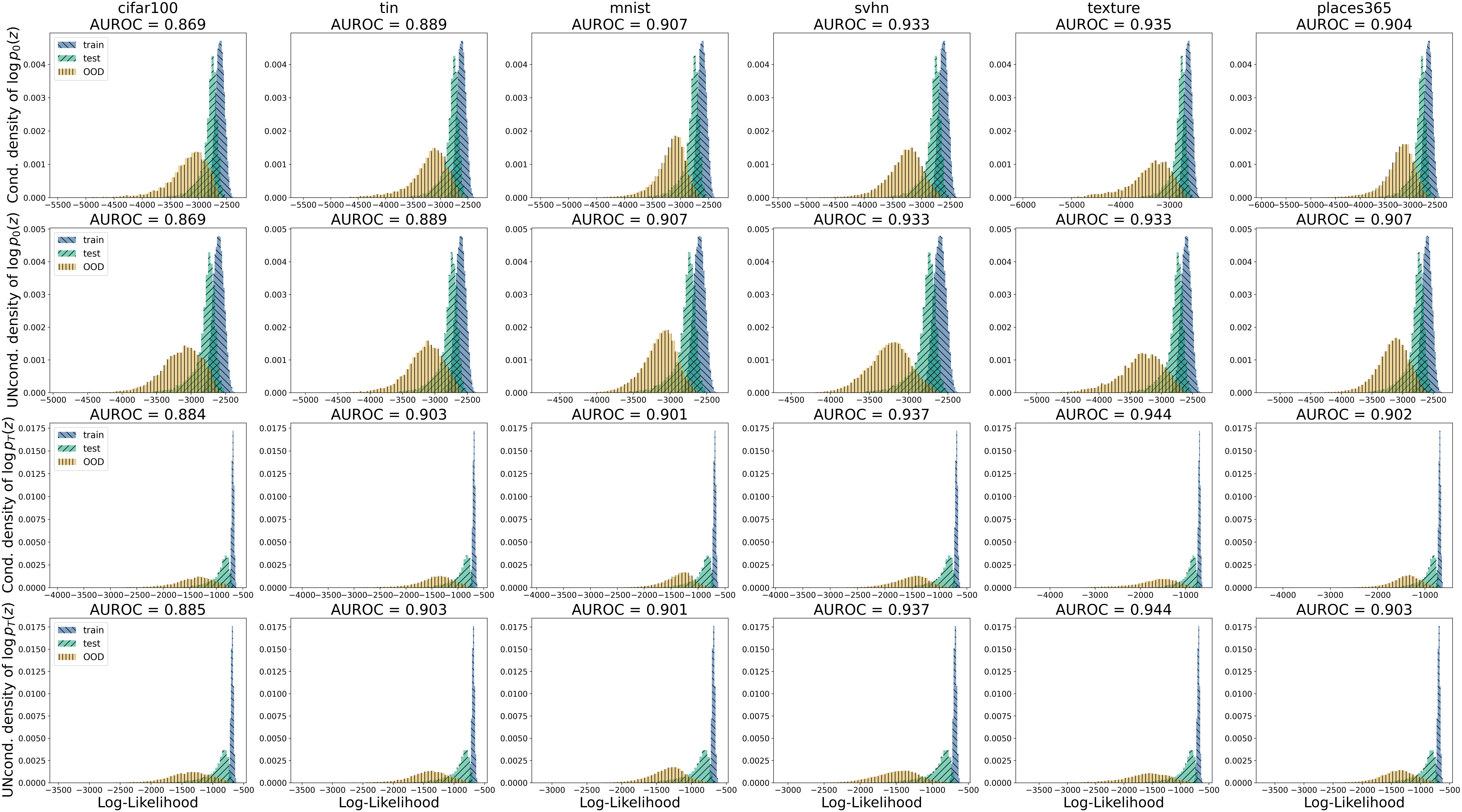}
    \caption{(InD: CIFAR-10) Comparison between using conditional vs. unconditional density with VDM trained for 2000 epochs. The resulting histograms (InD train (blue), test (green), and OOD (orange)) and AUROC scores are given for each OOD dataset and for both methods (EL and PL).}
\label{fig:dens_cif10_all}
\end{figure}

\subsection{CIFAR-100}

\begin{table}[H]
  \centering
  \caption{(InD: CIFAR-100). AUROC (\%) scores for our methods and baselines. Best in each column in \textbf{bold}, second best \underline{underlined}.}
  \label{tab:auroc_cifar100}
  \resizebox{0.9\columnwidth}{!}{%
  \begin{tabular}{lcccccccc}
    \toprule
    Method & CIFAR-10 & TIN & Near-OOD & MNIST & SVHN & Textures & Places365 & Far-OOD \\
    \midrule
    TempScale                     & \underline{79.02}   & 82.79   & \underline{80.9}    & 77.27   & 79.79   & 78.11   & \underline{79.80}   & 78.74    \\
    RMDS                          & 77.75   & 82.55   & 80.15   & 79.74   & \underline{84.89}   & 83.65   & \textbf{83.40}   & \textbf{82.92}    \\
    MLS                           & \textbf{79.21}   & \underline{82.90}   & \textbf{81.05}   & 78.91   & 81.65   & 78.39   & 79.75   & 79.67    \\
    VIM                           & 72.21   & 77.76   & 74.98   & \underline{81.89}   & 83.14   & \textbf{85.91}   & 75.85   & 81.70    \\
    KNN                           & 77.02   & \textbf{83.34}   & 80.18   & \textbf{82.36}   & 84.15   & 83.66   & 79.43   & \underline{82.40}    \\
    ASH                           & 76.48   & 79.92   & 78.20   & 77.23   & \textbf{85.60}   & 80.72   & 78.76   & 80.58    \\
    VDM $\log p_0(z)$   &73.12	&79.33	&76.23	&75.15	&81.40	&83.93	&77.03	&79.38\\
    VDM $\log p_T(z)$   &72.19&	79.55	&75.87	&73.46	&78.14	&\underline{83.99}	&76.62	&78.05\\
    TKDL                            & 77.69	&81.39	&79.54	&77.25&	75.48&	75.82	&78.49	&76.76\\

    \bottomrule
  \end{tabular}}
\end{table}

\begin{table}[H]
  \centering
  \caption{(InD: CIFAR-100) FPR@95 (\%) scores for our methods and baselines. Best in each column in \textbf{bold}, second best \underline{underlined}.}
  \label{tab:fpr_cifar100}
    \resizebox{0.9\columnwidth}{!}{%
  \begin{tabular}{lcccccccc}
    \toprule
    Method & CIFAR-10 & TIN & Near-OOD & MNIST & SVHN & Textures & Places365 & Far-OOD \\
    \midrule
    TempScale                     & \textbf{58.72}   & 50.26   & \textbf{54.49}   & 56.05   & 57.71   & 61.56   & \underline{56.46}   & 57.94    \\
    RMDS                          & 61.37   & \textbf{49.56}   & \underline{55.46}   & 52.05   & 51.65   & 53.99   & \textbf{53.57}   & \underline{52.81}    \\
    MLS                           & \underline{59.11}   & 51.83   & 55.47   & 52.95   & 53.90   & 62.39   & 57.68   & 56.73    \\
    VIM                           & 70.59   & 54.66   & 62.63   & \textbf{48.32}   & \underline{46.22}   & \textbf{46.86}   & 61.57   & \textbf{50.74}    \\
    KNN                           & 72.80   & \underline{49.65}   & 61.22   & \underline{48.58}   & 51.75   & 53.56   & 60.70   & 53.65    \\
    ASH                           & 68.06   & 63.35   & 65.71   & 66.58   & \textbf{46.00}   & 61.27   & 62.95   & 59.20    \\
    VDM $\log p_0(z)$  &69.56	&55.41	&62.48&	54.41	&48.53&	\underline{51.14}	&61.08	&53.79\\
    VDM $\log p_T(z)$ &72.89&	56.02	&64.46	&56.71	&54.10	&52.20	&63.40	&56.60\\
    TKDL                           &60.55	&50.96	&55.75	&55.52	&63.29	&64.24&	57.52	&60.14\\

    \bottomrule
  \end{tabular}}
\end{table}

\begin{figure}[H]
\centering
\includegraphics[width=0.99\linewidth]{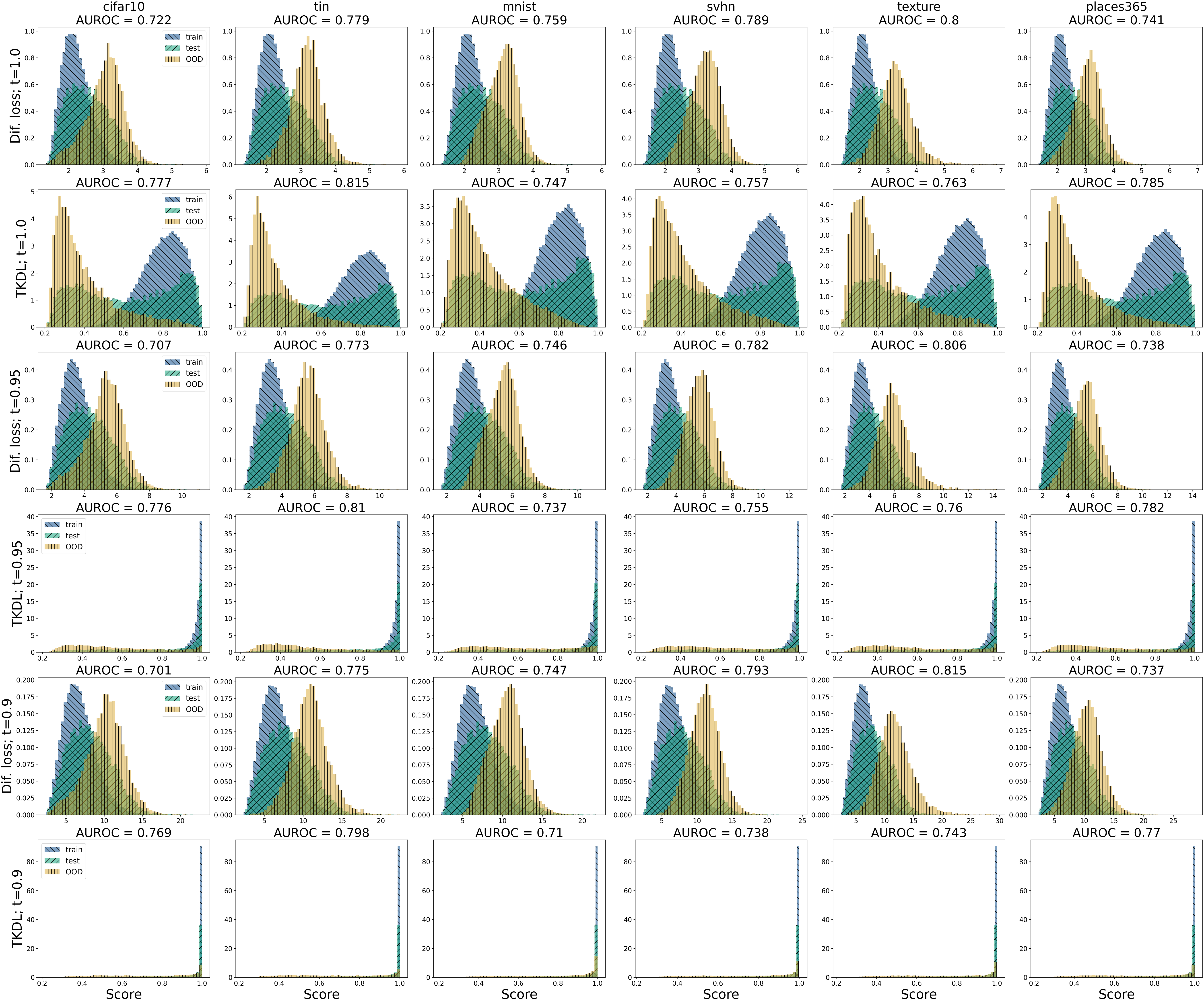}
    \caption{(InD: CIFAR-100) Comparison between using diffusion loss as an OOD score vs. our method TKDL with VDM trained for 2000 epochs. The resulting score histograms (InD train (blue), test (green), and OOD (orange)) and AUROC scores are given for each OOD dataset and for different diffusion timesteps (t=0.9, 0.95, 1.0) to demonstrate the relevance of the choice of timestep for both methods. }
\label{fig:basics_cif100_all}
\end{figure}

\begin{figure}[H]
\centering
\includegraphics[width=0.99\linewidth]{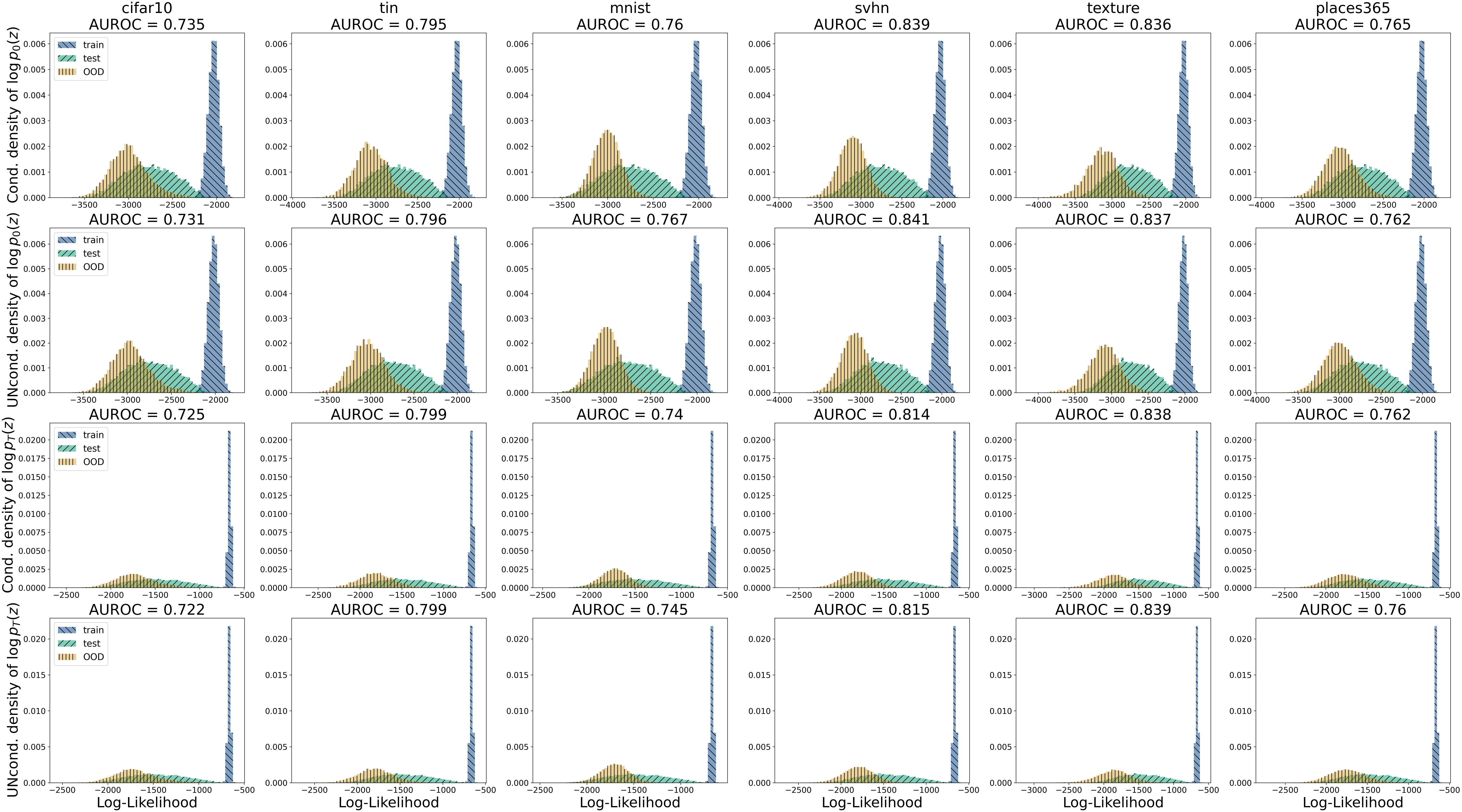}
    \caption{(InD: CIFAR-100) Comparison between using conditional vs. unconditional density with VDM trained for 2000 epochs. The resulting histograms (InD train (blue), test (green), and OOD (orange)) and AUROC scores are given for each OOD dataset and for both methods (EL and PL).}
\label{fig:dens_cif100_all}
\end{figure}

\subsection{ImageNet-200 Experiment results and figures}

\begin{table}[H]
  \centering
  \caption{(InD: ImageNet-200) AUROC (\%) scores for our methods and baselines. Best in each column in \textbf{bold}, second best \underline{underlined}.}
  \label{tab:auroc_imagenet200}
   \resizebox{0.9\columnwidth}{!}{%
  \begin{tabular}{lccccccc}
    \toprule
    Method & SSB-hard & NINCO & Near-OOD & iNaturalist & Textures & OpenImage-O & Far-OOD \\
    \midrule
    TempScale                     & \textbf{80.71}   & \textbf{86.67}   & \textbf{83.69}   & 93.39   & 89.24   & 89.84   & 90.82    \\
    RMDS                          & \underline{80.20}   & 84.94   & 82.57   & 90.64   & 86.77   & 86.77   & 88.06    \\
    MLS                           & 80.15   & 85.65   & \underline{82.90}   & 93.12   & 90.60   & 89.62   & 91.11    \\
    VIM                           & 74.04   & 83.32   & 78.68   & 90.96   & 94.61   & 88.20   & 91.26    \\
    KNN                           & 77.03   & \underline{86.10}   & 81.57   & \underline{93.99}   & \textbf{95.29}   & \underline{90.19}   & \underline{93.16}    \\
    ASH                           & 79.52   & 85.24   & 82.38   & \textbf{95.10}   & \underline{94.77}   & \textbf{91.82}   & \textbf{93.90}    \\
    VDM $\log p_0(z)$ & 53.65	&50.84	&52.24	&62.19	&71.38	&57.93	&63.83 \\
    VDM $\log p_T(z)$ & 56.74&	61.44&	59.09	&66.01	&81.05	&66.08	&71.05 \\
    TKDL &78.25&	83.60&	80.93&	88.09&	83.86&	85.20&	85.72 \\
    \bottomrule
  \end{tabular}}
\end{table}

\begin{table}[H]
  \centering
  \caption{(InD: ImageNet-200) FPR@95 (\%)  scores for our methods and baselines. Best in each column in \textbf{bold}, second best \underline{underlined}.}
  \label{tab:fpr_imagenet200}
   \resizebox{0.9\columnwidth}{!}{%
  \begin{tabular}{lccccccc}
    \toprule
    Method & SSB-hard & NINCO & Near-OOD & iNaturalist & Textures & OpenImage-O & Far-OOD \\
    \midrule
    TempScale                     & \underline{66.43}   & \underline{43.21}   & \underline{54.82}   & \underline{24.39}   & 43.57   & 34.04   & 34.00    \\
    RMDS                          & \textbf{65.91}   & \textbf{42.13}   & \textbf{54.02}   & 24.70   & 37.80   & 34.85   & 32.45    \\
    MLS                           & 69.64   & 49.87   & 59.76   & 25.09   & 41.25   & 35.76   & 34.03    \\
    VIM                           & 71.28   & 47.10   & 59.19   & 27.34   & \textbf{20.39}   & 33.86   & \textbf{27.20}    \\
    KNN                           & 73.71   & 46.64   & 60.18   & 24.46   & \underline{24.45}   & \textbf{32.90}   & \underline{27.27}    \\
    ASH                           & 72.14   & 57.63   & 64.89   & \textbf{22.49}   & 25.65   & \underline{33.72}   & 27.29    \\
    VDM $\log p_0(z)$ &92.80	&93.42	&93.11	&86.74	&84.84	&90.61&	87.40 \\
    VDM $\log p_T(z)$ & 92.35	&89.08	&90.71	&86.11	&75.16	&86.80 &	82.69 \\
    TKDL& 67.17	&45.98&	56.58	&32.84	&50.90	&40.87	&41.53\\
    \bottomrule
  \end{tabular}}
\end{table}

\begin{figure}[H]
\centering
\includegraphics[width=0.99\linewidth]{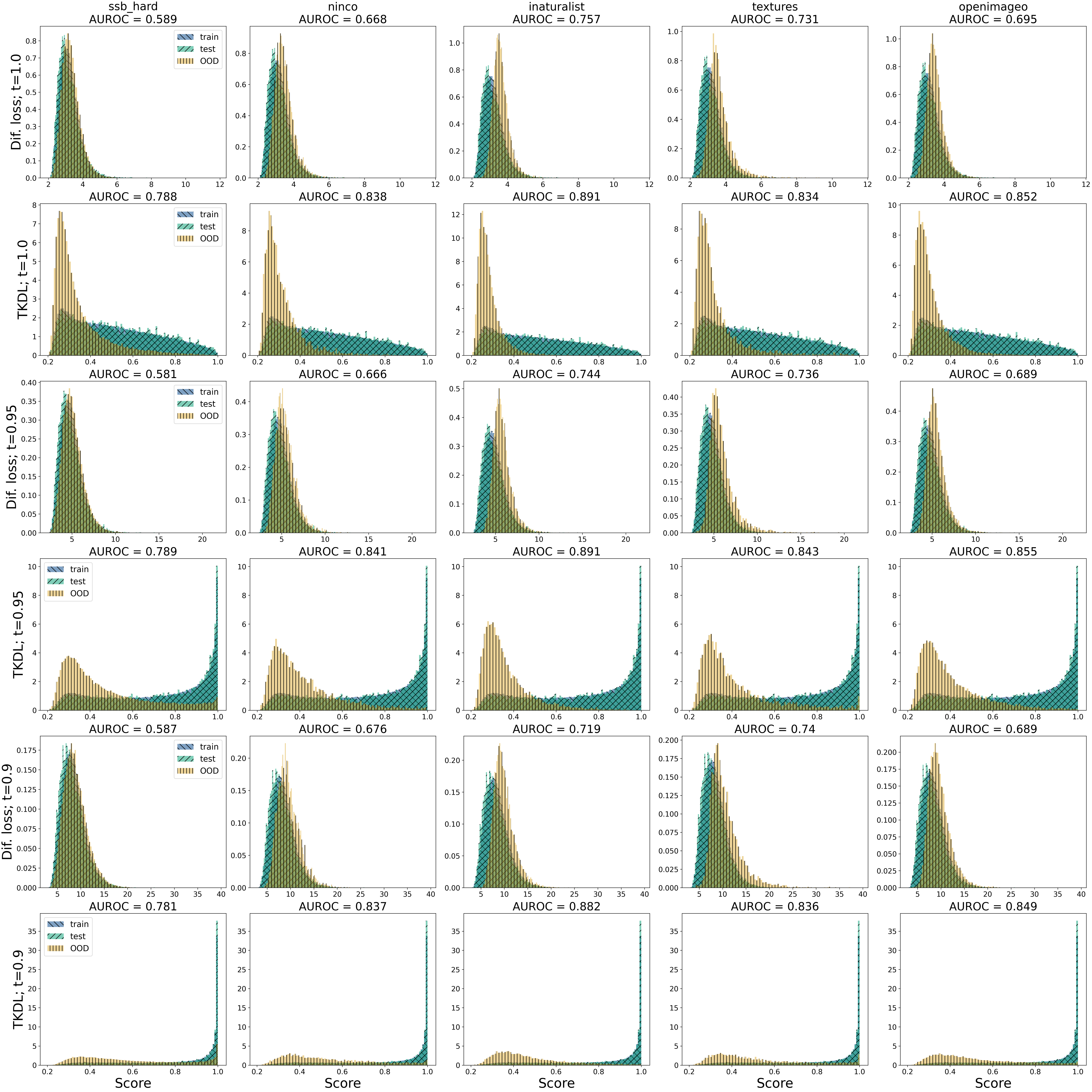}
    \caption{(InD: ImageNet-200) Comparison between using diffusion loss as an OOD score vs. our method TKDL with VDM trained for 2000 epochs. The resulting score histograms (InD train (blue), test (green), and OOD (orange)) and AUROC scores are given for each OOD dataset and for different diffusion timesteps (t=0.9, 0.95, 1.0) to demonstrate the relevance of the choice of timestep for both methods. }
\label{fig:basics_in200_all}
\end{figure}

\begin{figure}[H]
\centering
\includegraphics[width=0.99\linewidth]{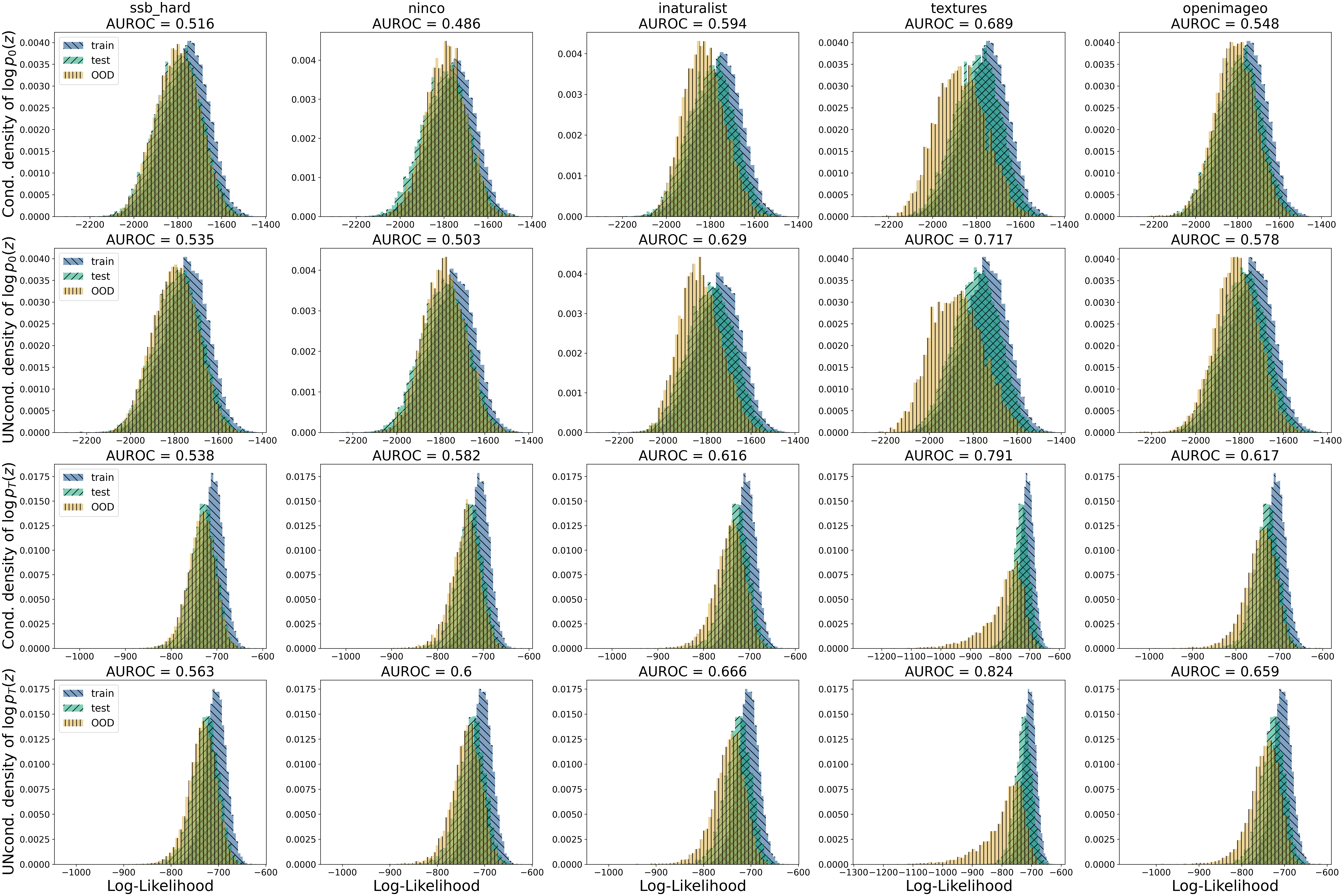}
    \caption{(InD: ImageNet-200) Comparison between using conditional vs. unconditional density with VDM trained for 2000 epochs. The resulting histograms (InD train (blue), test (green), and OOD (orange)) and AUROC scores are given for each OOD dataset and for both methods (EL and PL).}
\label{fig:dens_in200_all}
\end{figure}

\section{Code}\label{code}
The code of our work is accessible from GitHub \url{https://github.com/joonasrooben/vldm_ood}.
Parts of our code were used from Github repositories such as: "Variational Diffusion Models in Easy PyTorch"\footnote{\url{https://github.com/myscience/variational-diffusion/tree/main}}, "OpenOOD"\footnote{\url{https://github.com/Jingkang50/OpenOOD/tree/main}}, "Score-Based Generative Modeling through Stochastic Differential Equations"\footnote{\url{https://github.com/yang-song/score_sde}}.